\pdfoutput=1
\documentclass[11pt]{article}
\usepackage{acl}
\usepackage{times}
\usepackage{latexsym}
\usepackage[T1]{fontenc}
\usepackage[utf8]{inputenc}
\usepackage{microtype}
\usepackage{inconsolata}
\usepackage{graphicx}
\usepackage{hyperref}
\usepackage{url}
\usepackage{algorithm}
\usepackage{algorithmic}
\usepackage{booktabs} 
\usepackage{multirow} 
\usepackage{lineno}
\usepackage{amsmath}

\title{SAUP: Situation Awareness Uncertainty Propagation on LLM Agent}

\author{
\textbf{Qiwei Zhao}$^{1}$, \textbf{Xujiang Zhao}$^{2}$, \textbf{Yanchi Liu}$^{2}$, \textbf{Wei Cheng}$^{2}$, \textbf{Yiyou Sun}$^{2}$, \\
\textbf{Mika Oishi}$^{3}$, \textbf{Takao Osaki}$^{3}$, \textbf{Katsushi Matsuda}$^{3}$, \textbf{Huaxiu Yao}$^{1}$, \textbf{Haifeng Chen}$^{2}$ \\
$^{1}$University of North Carolina at Chapel Hill, $^{2}$NEC Labs America, $^{3}$NEC Corporation \\
\texttt{qiwei@cs.unc.edu, xuzhao@nec-labs.com} \\
}

\begin{document}
\maketitle
\begin{abstract}
Large language models (LLMs) integrated into multistep agent systems enable complex decision-making processes across various applications. However, their outputs often lack reliability, making uncertainty estimation crucial. Existing uncertainty estimation methods primarily focus on final-step outputs, which fail to account for cumulative uncertainty over the multistep decision-making process and the dynamic interactions between agents and their environments. To address these limitations, we propose SAUP (Situation Awareness Uncertainty Propagation), a novel framework that propagates uncertainty through each step of an LLM-based agent's reasoning process. SAUP incorporates situational awareness by assigning situational weights to each step's uncertainty during the propagation. Our method, compatible with various one-step uncertainty estimation techniques, provides a comprehensive and accurate uncertainty measure. Extensive experiments on benchmark datasets demonstrate that SAUP significantly outperforms existing state-of-the-art methods, achieving up to 20\% improvement in AUROC.
\end{abstract}
\section{Introduction}

Large language models (LLMs) \cite{minaee2024large} have demonstrated remarkable capabilities and, when integrated into agent systems \cite{wang2024survey}, enable complex decision-making processes and broader applications. However, while LLM-based agents are increasingly effective, their outputs are not always reliable, which can lead to significant issues, particularly in high-stakes environments such as healthcare or autonomous systems. This makes uncertainty estimation critical, as it evaluates the reliability of an agent's decisions and outputs \cite{chang2024survey, raiaan2024review}.  Understanding and quantifying uncertainty is essential because it offers insight into potential system failures, providing a safeguard for sensitive applications. Current methods for estimating uncertainty in LLM-based agents remain limited. For example, UALA \cite{han2024towards} proposes a one-step uncertainty measurement to estimate the uncertainty of the final step before the agent provides an answer.

\begin{figure}
\centering
\includegraphics[ width=0.45\textwidth]{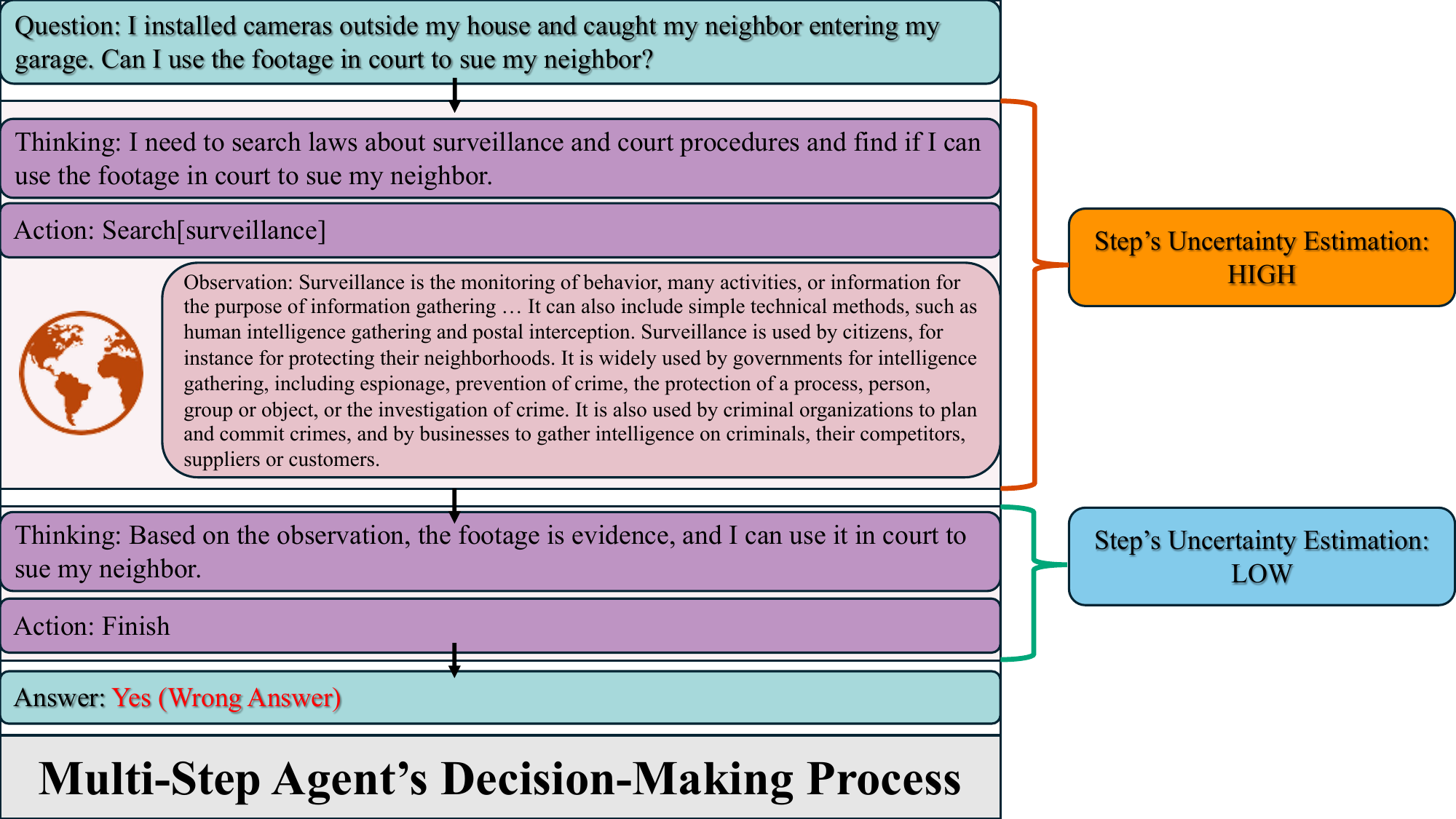}
\caption{The overall uncertainty of an agent based on large language models (LLMs) can arise from two primary sources: a) \textit{Uncertainty Across All Steps}: Encompassing both intermediate and final steps; and b) \textit{The Agent's Situational Context}: Including the quality of its interaction with the environment and deviations from the optimal logical path. \textit{In this example:} A user installs security cameras and captures footage of a neighbor entering her garage without permission. She asks an LLM-based agent whether this footage can be used in court. The agent first searches for information on surveillance laws, identifying a definition related to intelligence and crime prevention. It then concludes that the footage qualifies as evidence, based on this research. However, the agent overlooks critical legal factors such as privacy laws and rules on admissibility of evidence, leading to an incorrect conclusion.}
\label{image:motivate}
\end{figure}

A key challenge is that uncertainty accumulates over time in multi-step processes, rather than in isolated actions, and is further exacerbated in dynamic environments where external factors are uncontrollable. These interactions can significantly impact the system's overall uncertainty. Therefore, robust methods that account for various information sources and interaction complexities are necessary to accurately capture the uncertainty across an agent's entire decision-making process. As illustrated in Figure \ref{image:motivate}, in sensitive contexts, solely observing the final step's uncertainty may lead to overconfidence in the outcome, resulting in adverse consequences and highlighting the importance of considering intermediate uncertainties and the quality of interaction between the agent and its environment.

To estimate LLM uncertainty, previous approaches focus mainly on the variance of the final step's output at the token, sentence, or semantic level. Predictive entropy \cite{gal2016dropout,gal2017deep}, initially used in image data, was extended to language models to predict uncertainty in output tokens \cite{xiao2021hallucination}. Although likelihood can also indicate uncertainty, \cite{malinin2020uncertainty} introduces normalized entropy, accounting for the output length. \cite{kuhn2023semantic} proposes semantic entropy, incorporating linguistic invariances within shared meanings. \cite{kadavath2022language, yin2023large} explore self-assessment by LLMs to estimate uncertainty. However, these methods, designed for traditional one-step QA, do not directly apply to LLM agents.\textit{ They face two key issues: first, they only consider the final step’s uncertainty, ignoring the accumulation of uncertainty throughout the process; second, they overlook the reasoning process of LLM agents, which is critical in multi-step decision-making and the agent's interaction with its environment.}

To address the challenges of uncertainty in multi-step processes within complex environments, we introduce \textbf{SAUP} (Situation-Awareness Uncertainty Propagation). SAUP comprehensively estimates uncertainty in LLM-based agents by propagating uncertainty through the multi-step reasoning and decision-making process. It builds upon frameworks like ReACT \cite{yao2022react}, which integrates LLMs' reasoning into problem-solving by decomposing tasks into thinking, acting, and observing steps. SAUP propagates uncertainties from the initial stages to the final step and aggregates them using a situation-weighting scheme, where each step's uncertainty is weighted based on the agent's situation, progress, and observation quality. Since directly measuring an agent's situation is challenging, we design effective surrogates that are adaptable to various scenarios. 

The primary contribution of this paper can be summarized as follows: \underline{Firstly}, We propose SAUP, a simple yet effective pipeline for providing comprehensive situation-aware uncertainty estimation in multi-step agents within complex environments. Unlike existing single-step uncertainty estimation methods, SAUP accounts for the agent's situational context throughout problem-solving, rather than focusing solely on the final step. \underline{Secondly}, To estimate the agent's unobservable situation, we introduce surrogate methods, which excel in estimating situational uncertainty and offer potential applications in related fields. \underline{Lastly}, We evaluate SAUP on benchmark datasets such as HotpotQA \cite{yang2018hotpotqa}, StrategyQA \cite{geva2021did}, and MMLU \cite{hendrycks2020measuring}. SAUP outperforms state-of-the-art methods, achieving up to a 20\% improvement in AUROC, demonstrating its effectiveness. 

\section{Related Works}
\subsection{LLM-based Agent} 
The reasoning capabilities of LLMs have prompted researchers to explore their use as the core of agent reasoning. Nakano et al. \cite{nakano2021webgpt} made an early attempt to employ LLMs as agents with web search and information retrieval capabilities, transitioning LLMs from passive tools to proactive agents interacting with complex environments. Subsequent works \cite{wang2021codet5, chen2021evaluating} explored LLMs in code generation for software development. Yao et al. \cite{yao2022react} introduced the ReAct pipeline, utilizing LLMs for decision-making where agents retrieve external information before making decisions. This framework, mirroring human decision-making, became foundational for decision-making agents, inspiring improvements by Shinn et al. \cite{shinn2023reflexion} and Renze et al. \cite{renze2024self} through self-reflection. Li et al. \cite{li2023camel} proposed CAMEL, which expanded the framework to enable communication between agents, fostering collaboration. Similarly, AutoGen \cite{wu2023autogenenablingnextgenllm} allows agents to converse and collaborate with customizable interactions in natural language and code. To further enhance decision-making, Qiao et al. \cite{qiao2023making} incorporated tool-based monitoring to refine agent behaviors.

\subsection{Uncertainty in Large Language Models} 
LLMs dominate numerous fields, including as agents \cite{zhao2023survey, xi2023rise}, but targeted uncertainty estimation methods for LLM-based agents remain unexplored. Existing techniques focus on one-step output uncertainty, originating from traditional language models, such as methods to improve model calibration \cite{xiao2019quantifying, xiao2021hallucination, jiang2021can}. Token-level uncertainty estimation in "white-box" LLMs \cite{malinin2020uncertainty, fomicheva2020unsupervised, darrin2022rainproof, duan2024shifting} has advanced, with Kuhn et al. \cite{kuhn2023semantic} introducing semantic equivalence into these calculations. Additionally, self-estimation of uncertainty in both "white-box" and "black-box" LLMs, accessed via APIs, has been explored \cite{kadavath2022language, yin2023large, chen2024hytrel}. These methods focus on one-step uncertainty estimation, which can be integrated into the SAUP framework as the backbone for uncertainty assessments.

\section{SAUP: Situational Awareness Uncertainty Propagation}
\begin{figure*}
\centering
\includegraphics[ width=0.85\textwidth]{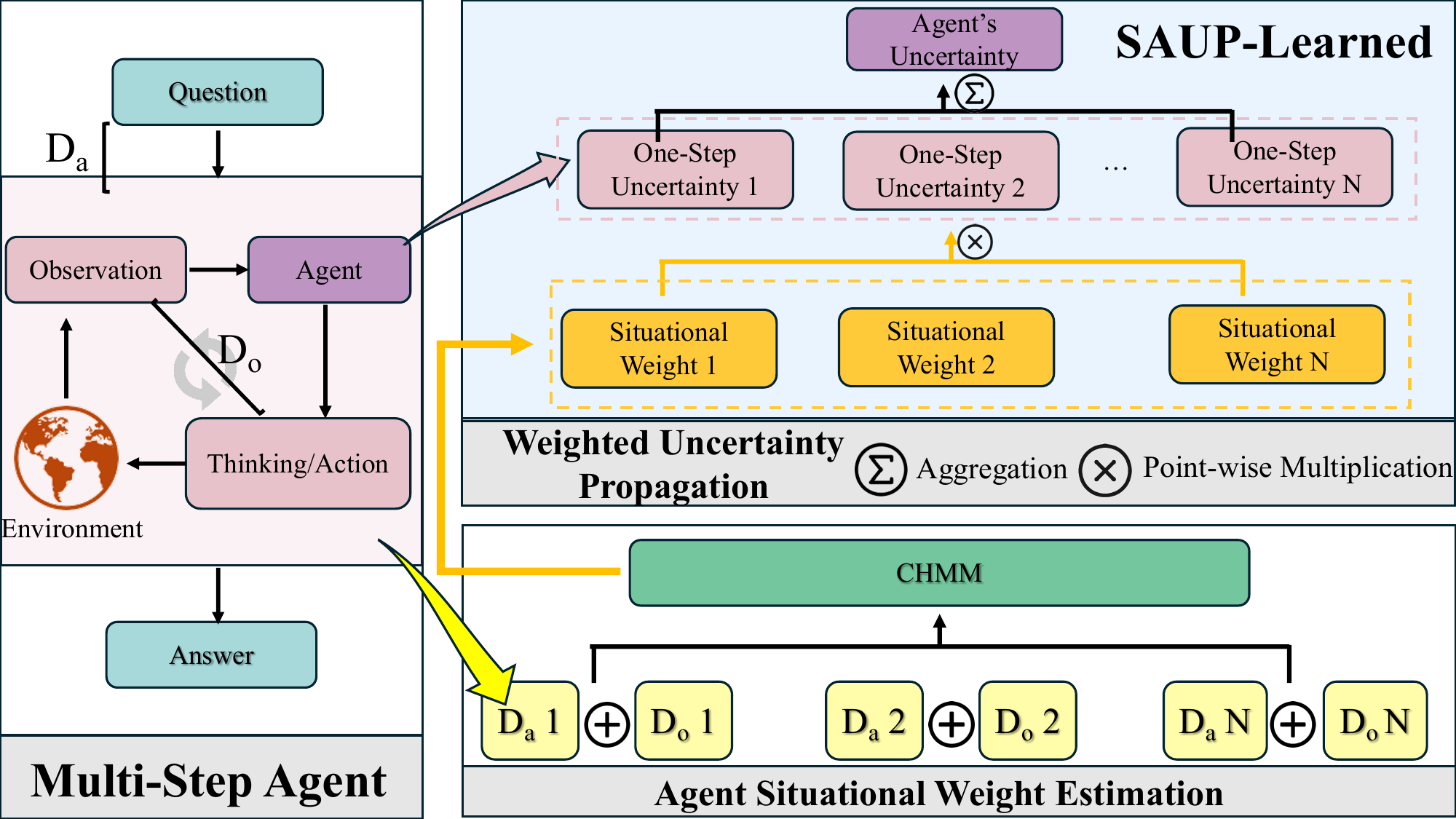}
\caption{Overview of our proposed SAUP, which is illustrated in three parts.  \textit{Left} depicts the general pipeline of LLM-based multi-step agents interacting with their environment. This process typically involves three behaviors: thinking, action, and observation. The  $D_a$ represents the distance between the question and the combination of thinking, action, and observation, whereas $D_o$ denotes the distance between the observation and the thinking/action. \textit{Bottom Right} illustrates the agent's situational weight estimation. Here, we employ a Hidden Markov Model (HMM) to estimate the situational weight based on the distances $D_a$ and $D_o$. \textit{Top Right} shows the process of weighted uncertainty propagation, where we aggregate the one-step uncertainty and the corresponding situational weight to derive the agent's overall uncertainty.}
\label{image:main}
\end{figure*}

We propose our pipeline, SAUP, with the goal of accurately estimating the overall agent's uncertainty by comprehensively considering the uncertainty at each step and the corresponding situational weights, as described in Figure \ref{image:main}. In the following sections, we delve into the details, elucidating how we aggregate the uncertainty from each step and estimate the corresponding situational weights.

\subsection{Weighted Uncertainty Propagation}

\textbf{Uncertainty Propagation.} As depicted in the \textit{left} part of Figure \ref{image:main}, for each step \(i\), the agent provides the thinking/action with the corresponding uncertainty \(U_i\) based on the previous state \(Z_{i-1}\) and the question \(Q\). Considering only the uncertainty of the last step as the overall uncertainty \(U_{agent}\) is unreasonable and not comprehensive. Instead, we should comprehensively consider and propagate the uncertainties of all steps. The simplest example is using an arithmetic mean of the uncertainty across the steps before the agent gives the final answer. For robustness against outliers, accurate reflection of central tendency, and consistency in proportional changes, the geometric mean or Root Mean Square (RMS) can be a better choice compared to the arithmetic mean.

\textbf{Situational Uncertainty Weights.} Based on the intuitive logic of information flow and experimental observations, we have identified that the contribution of uncertainty at different steps to the overall agent uncertainty is not uniform. Therefore, in addition to the uniform aggregation scheme introduced earlier, it is essential to design a more comprehensive weighting aggregation scheme for overall uncertainty, tailored to the characteristics of the agent.

During the process of obtaining the final answer, the LLM-based agent produces uncertainty. We refer to the contribution of the current step's uncertainty to the overall uncertainty, due to the agent's situation, as the situational weights. Situational weights are determined by factors, such as deviations from the appropriate logical path and the quality of interactions between the agent and the environment, which influence the correctness of the final answer. These situational weights are variable during the agent's problem-solving process and its interaction with the environment. Assume that the uncertainty at step \(i\) is \(U_i\) and the corresponding situational weight is \(W_i\), the formula of weighted uncertainty propagation is:

\begin{equation}
\label{equ: aggregate}
U_{\text{agent}} = \sqrt{\frac{1}{N} \sum_{i=1}^{N} \left(  (W_i U_i)^2  \right)}
\end{equation}

Here we choose the RMS as the propagation method. In the practical application of SAUP, besides the above linear term, we also utilize an extra logical term for numerical stability. We designed the SAUP formula based on the following considerations. \underline{First}, SAUP relies on a comprehensive consideration of all steps of the agent based on propagation. \underline{Second}, by introducing situational weights for the uncertainty of different steps, SAUP allows for a more complete assessment of the impact of specific steps on the overall uncertainty of the agent. In the following section \ref{sec: Step Uncertainty Estimation} and  \ref{sec: Agent Situational Estimation}, we will introduce the method for calculating the uncertainty $U_i$ and  the situational weight \(W_i\) corresponding to each step.

\subsection{Step Uncertainty Estimation}
\label{sec: Step Uncertainty Estimation}
From equation \ref{equ: aggregate}, we can see that essentially, our SAUP is compatible with all single-step uncertainty estimation methods applicable to various scenarios, including but not limited to the ones we mentioned. SAUP is built upon these one-step methods.

In the practical implementation, we utilize the normalized entropy \cite{malinin2020uncertainty}, with some modifications to adapt it to the characteristics of the React Agent pipeline. This choice is based on the consideration that normalized entropy has broad applicability. It can not only be applied to open-source LLMs, such as LLAMA, where complete logits of the output are accessible, but can also be utilized with LLMs that are accessible only via API, such as the CHATGPT series. In addition, it is computationally efficient and demonstrates strong predictive performance for single-step uncertainty estimation. 

For step \(n\) and question $Q$, we denote the agent's thinking as \(T_n\), and the corresponding action as \(A_n\). The observation \(O_n\) is the information gained from the environment through the action \(A_n\). Let the LLM be denoted as \(L_\theta\), and the trajectory of the previous \(n-1\) steps as \(Z_{n-1}\), where \(Z_{n-1} = \{(A_1, T_1, O_1), \ldots, (A_{n-1}, T_{n-1}, O_{n-1})\}\). The LLM will output the thinking \(T_n\) and the action \(A_n\) together as:

\begin{equation}
    (T_{n}, A_{n}) = L_{\theta}(Q, Z_{n-1}) 
\end{equation}

Suppose \(T_n\) consists of the first \(N\) tokens, while \(A_n\) consists of the following \(M\) tokens. The uncertainty \(U_n\) for step \(n\) is computed by the following equation:

\begin{equation}
\begin{aligned}
&U_n = \frac{1}{N+M} \prod_{i \leq N+M} p(t_i \mid t_0, \ldots, t_{i-1} ; \theta)\\
&= \frac{1}{N+M} \sum_{i \leq N+M} log\ p(t_i \mid t_0, \ldots, t_{i-1} ; \theta)
\end{aligned}
\label{equ:LLM_LL}
\end{equation}

Where \( p(t_i \mid t_0, \ldots, t_{i-1} ; \theta) \) is the token probability of \(\text{token i} \), given the previous \(\text{token 0} , \dots, \text{token i-1}\), and the parameters \(\theta\) of the LLM \(L\).

\subsection{Agent Uncertainty Estimation}
\label{sec: Agent Situational Estimation}

Assigning weights $W_i$ to each step's uncertainty $U_i$ in a multi-step reasoning process is crucial for accurate overall uncertainty estimation. In LLM-based agents, effective reasoning significantly influences decision-making. However, these agents may exhibit overconfidence, making it essential to evaluate their situational state properly. Since the situational state is not directly observable, surrogate measures are used to approximate it. One idea is to assign greater weight to steps closer to the final answer or to measure deviation from an ideal trajectory. While these approaches have merit, they do not fully capture the agent's true situational state.

To address this limitation, we propose learning-based surrogates that target the agent's hidden situations. Among these, the Hidden Markov Model (HMM) Distance Surrogate (SAUP-HMMD) learns step transitions and assigns weights based on hidden state estimations. HMMs stand out for their minimal data requirements and computational efficiency, making HMMD the preferred surrogate in cases where training data is limited. In contrast, more complex sequence-to-sequence (S2S) models like Long Short-Term Memory networks (LSTMs) and Transformers capture intricate temporal dependencies but require significantly more training data and time. While any S2S model can theoretically be used for this task, the choice largely depends on the size of the training dataset, with HMM being the default choice when the dataset is small. A more detailed analysis and comparison of these learned surrogates are provided in Section \ref{sec: Dissection}.

A Hidden Markov Model (HMM, \cite{baum1966statistical}) estimates hidden states based on observable ones, assuming regular transitions between hidden states. An HMM is defined by the number of hidden states \( N \) and observable states \( M \), with hidden states \( S_{hmm} = \{S_{hmm_1}, \ldots, S_{hmm_N}\} \) and observable states \( O_{hmm} = \{O_{hmm_1}, \ldots, O_{hmm_M}\} \). The state transition probability matrix \( A = [a_{ij}] \) represents \( P(S_{hmm_j} \mid S_{hmm_i}) \), and the observation probability matrix \( B = [b_{jk}] \) represents \( P(O_{hmm_k} \mid S_{hmm_j}) \). The initial state distribution \( \pi = [\pi_i] \) defines the probability of starting in state \( S_{hmm_i} \). In Continuous Hidden Markov Models (CHMM), observations are modeled by continuous probability density functions, typically Gaussian Mixture Models (GMMs). We adopt CHMMs as the backbone model for HMMD. The CHMM defines three discrete hidden states: correct trajectory, moderately deviated trajectory, and highly deviated trajectory. The observable states are continuous features, represented by the two-feature plain distance \( (D_a, D_o) \). The specific method for calculating the plain distance between A and B, denoted as $dis(A, B)$, utilizes a pre-trained RoBERTa \cite{liu2019roberta} model, fine-tuned with the SQuAD v2 \cite{rajpurkar2018know} dataset.  The inverse of the score obtained from this model is used as the plain distance. Using training examples, we calculate \( (D_a, D_o) \) and annotate the hidden states. And The CHMM is trained using the Baum-Welch algorithm \cite{baum1970maximization}, transforming the two-feature plain distance into a more accurate surrogate for the agent’s situational awareness.

The SAUP algorithm employs different surrogate configurations. We illustrate the SAUP using distance as the surrogate in Algorithm \ref{alg:SAUP}. \underline{Initially}, uncertainty $U_n$ is computed for step $n$, along with the corresponding distances $D_{a_n}$ and $D_{o_n}$. This is repeated for $N$ steps. \underline{Subsequently}, based on the surrogate choice, either plain or HMM-based, the situational weights $W_n$ are determined. \underline{Finally}, the uncertainties $U$ and weights $W$ are aggregated to estimate the agent’s overall uncertainty $U_{agent}$.

\begin{algorithm}[th]
\caption{Situational Awareness Uncertainty Propagation (SAUP) }
\label{alg:SAUP}
\begin{algorithmic}
    \STATE Initialize the $N$-Step LLM-based Agent $L_\theta$ with the problem $Q$, and the $Z_n = \{(A_1, T_1, O_1), (A_2, T_2, O_2), \ldots, (A_{n}, T_{n}, O_{n})\}$, the List $D_L$ to store the distance, the trained CHMM model $H$,  the distance calculate method $Dis()$ from the section \ref{sec: Agent Situational Estimation}, the single-step uncertainty method $F_U$, and the situation awareness uncertainty propagation function SAUP(), defined as the equation\ref{equ: aggregate}.
    \FOR{step $n$ in the problem solving process}
        \STATE The Uncertainty for current step $U_n \leftarrow F_U(L_\theta, Z_n)$
        \STATE Distance $D_{a_n} \leftarrow Dis(Z_n,Q)$ 
        \STATE Distance $D_{o_n} \leftarrow Dis(A_n,O_n)$        

        \IF{using the \textit{HMM-Distance} as the surrogates}
            \STATE Add the $(D_{a_n} + D_{o_n})$ into the $D_L$
        \ELSE
            \STATE Using the \textit{Plain-Distance} as the surrogates
            \STATE $W_n \leftarrow D_{a_n} + D_{o_n}$ 
        \ENDIF
    \ENDFOR
    \IF{using the \textit{HMM-Distance} as the surrogates}
    \STATE $(W_1, W_2, \ldots, W_N) \leftarrow H(D_L) = H((D_{a_1} + D_{o_1}), \ldots,(D_{a_N} + D_{o_N}) )$
    \ENDIF
    \STATE The Uncertainty for the agent $U_{agent} \leftarrow SAUP((U_1,W_1),(U_2, W_2), \ldots, (U_N,W_N))$
    \RETURN Situational Awareness Agent Uncertainty $\mathbf{U_{agent}}$
\end{algorithmic}
\end{algorithm}

\section{Experiments}
In this section, we evaluate the performance of SAUP, aiming to answer the following questions: \textbf{Q1}: Does SAUP outperform previous state-of-the-art approaches for uncertainty estimation? \textbf{Q2}: Given the comprehensive process of Uncertainty Propagation, does SAUP provide more accurate uncertainty estimation compared to single-step methods? \textbf{Q3}: Are the situational weights for specific steps effective in improving overall uncertainty estimation? Since obtaining precise situational weights is impractical, we designed surrogates, including distance-based and position-based methods. Are these surrogates reliable for accurately assessing the agent's current situation?

\subsection{Experimental Setup}
\label{sec: Re_agent}

\textbf{LLM-based Agent Framework.} Our experiments focus on evaluating SAUP's ability to improve uncertainty estimation for multi-step LLM-based agents. While various multi-step agents follow different pipeline designs, they generally adhere to the thinking-acting-observation workflow. We chose the React \cite{yao2022react} framework, a widely-used agent model, for its alignment with this workflow.

\textbf{Backbone LLMs.} We selected two categories of LLMs for the React agents: the open-source LLAMA3 \cite{dubey2024llama} series (8B and 70B models) with entropy access, and GPT-4o \cite{achiam2023gpt} (available via API), which restricts internal information. This selection ensures broad coverage of real-world scenarios.

\textbf{Dataset and Task.} We evaluated three challenging agent-based QA tasks. The first, \textbf{HotpotQA} \cite{yang2018hotpotqa}, focuses on multi-hop QA with diverse free-form answers. We randomly sampled 2,000 questions from the development set, assessed by both human evaluators and ChatGPT. The second, \textbf{MMLU} \cite{hendrycks2020measuring}, involves multiple-choice questions across diverse fields like law and mathematics. Ten questions were sampled per subtask from the test set. Lastly, \textbf{StrategyQA} \cite{geva2021did} requires implicit reasoning, evaluated with true/false questions from its development set (229 questions).

\textbf{Environment for External Information.} LLM-based agents often need external sources to solve these tasks. For HotpotQA and StrategyQA, we provided access to the Wikipedia API, which retrieves relevant entity-based information. For MMLU, we used SerpAPI \cite{serpapi} for structured Google search results.

\textbf{Baselines.} We evaluated SAUP against several uncertainty estimation methods. For entropy-based approaches, we used predictive and semantic entropy \cite{xiao2019quantifying, kuhn2023semantic}. Likelihood-based methods \cite{malinin2020uncertainty} included plain likelihood and normalized entropy, the latter accounting for token length. We also implemented P(True) \cite{kadavath2022language, yin2023large}, which prompts agents to self-assess their confidence.

\textbf{Evaluation Metrics.} We used AUROC \cite{bradley1997use} to measure the ability of uncertainty methods to distinguish between correct and incorrect responses. Higher AUROC values indicate better differentiation, with a perfect score of 1 representing complete distinction and 0.5 representing random chance.

\subsection{Superior Discriminative Performance of SAUP}

\begin{table*}[ht]
\small
\begin{center}
\caption{Results for SAUP. The best results and second best results are \textbf{bold} and \underline{underlined}, respectively.}
\label{tab:results_all}
\resizebox{2.0\columnwidth}{!}{\setlength{\tabcolsep}{1.5mm}{
\begin{tabular}{cc|c|c|c|c|c|c|c|c|c}
\toprule
\multicolumn{2}{c|}{}& \multicolumn{3}{c|}{\textbf{HotpotQA}} & \multicolumn{3}{c|}{\textbf{MMLU}}& \multicolumn{3}{c}{\textbf{StrategyQA}} \\
\cmidrule{3-11}
\multicolumn{2}{c|}{\raisebox{2.5ex}[0pt]{\textbf{Method}}}&LLAMA3 8B & LLAMA3 70B & GPT4O & LLAMA3 8B & LLAMA3 70B & GPT4-O & LLAMA3 8B & LLAMA3 70B & GPT4-O \\\midrule
&Predictive Entropy & 0.631 & 0.617 & N.A. &0.531 &0.585 & N.A. &0.542 &0.589& N.A.\\
& Likelihood & 0.653 & 0.622 & 0.764 & 0.550 &0.592 &\underline{0.610}&0.525&0.591 & 0.641\\
& Normalised Entropy & 0.664 & 0.635 & \underline{0.772} & \underline{0.555} &0.579 &0.607&0.554 &0.557 & \underline{0.710} \\
& P(True) & 0.601 & 0.618 & 0.749 & 0.528  &0.560 &0.588&0.533 &0.577 & 0.689 \\
& Semantic Entropy & \underline{0.702} & \underline{0.669} & N.A. & 0.548 &\underline{0.605} &N.A.&\underline{0.599} &\underline{0.610} & N.A. \\
\midrule
& \textbf{SAUP-Learned} & \textbf{0.771} & \textbf{0.755} & \textbf{0.778}  & \textbf{0.669} &\textbf{0.638} & \textbf{0.626}&\textbf{0.787} &\textbf{0.783} & \textbf{0.809}\\
\bottomrule
\end{tabular}}}
\end{center}
\vspace{-1.5em}
\end{table*}

\begin{table*}[ht]
\small
\begin{center}
\caption{Results for SAUP with various Surrogates. The best results and second best results are \textbf{bold} and \underline{underlined}, respectively.}
\label{tab:results_surrogate}
\resizebox{2.0\columnwidth}{!}{\setlength{\tabcolsep}{1.5mm}{
\begin{tabular}{cc|c|c|c|c|c|c|c|c|c}
\toprule
\multicolumn{2}{c|}{}& \multicolumn{3}{c|}{\textbf{HotpotQA}} & \multicolumn{3}{c|}{\textbf{MMLU}}& \multicolumn{3}{c}{\textbf{StrategyQA}} \\
\cmidrule{3-11}
\multicolumn{2}{c|}{\raisebox{2.5ex}[0pt]{\textbf{Method}}}&LLAMA3 8B & LLAMA3 70B & GPT4O & LLAMA3 8B & LLAMA3 70B & GPT4-O & LLAMA3 8B & LLAMA3 70B & GPT4-O \\
\midrule 
& SAUP-P & 0.723 & 0.739 & \textbf{0.797} & 0.634& \underline{0.636} & 0.614&0.668 &0.641 & 0.734\\
& SAUP-D & \underline{0.762} & 0.726 & 0.773 & \underline{0.660}& 0.619 & \underline{0.624}&\underline{0.755} &\textbf{0.809} & \underline{0.806}\\
& SAUP-PD & 0.759 & \underline{0.745} & 0.782 & 0.651&0.625 & 0.619&0.732 &0.756 & 0.785\\
\midrule
& \textbf{SAUP-HMMD(Learned)} & \textbf{0.771} & \textbf{0.755} & \underline{0.778}  & \textbf{0.669} &\textbf{0.638} & \textbf{0.626}&\textbf{0.787} &\underline{0.783} & \textbf{0.809}\\
\bottomrule
\end{tabular}}}
\end{center}
\vspace{-1.5em}
\end{table*}

In this section, we compare the performance of various uncertainty measurement methods in distinguishing whether an LLM-based agent's final response to QA questions is correct or incorrect. The evaluation process consists of the following steps: (1) The LLM-based agent, using the ReACT framework, answers the QA questions; (2) Multiple versions of our proposed SAUP method, along with other baseline uncertainty estimation methods, compute an uncertainty score for each agent's response; (3) Each response is assessed for correctness, assigning a value of 0 if the answer is correct and 1 if incorrect; (4) We calculate the AUROC based on the accuracy of these classifications and the corresponding uncertainty scores. Ideally, incorrect answers should correlate with higher uncertainty scores.

We employed several state-of-the-art LLMs, including \textit{\{LLAMA3 8B, LLAMA3 70B, GPT4O\}}, and conducted evaluations on challenging datasets, namely \textit{\{StrategyQA, MMLU, HotpotQA\}}. Table \ref{tab:results_all} presents the results, demonstrating that our SAUP method, consistently achieves higher AUROC scores across all datasets compared to state-of-the-art methods. These findings indicate that SAUP offers superior performance in distinguishing between correct and incorrect agent responses based on uncertainty estimation, leading to important conclusions.

\begin{table*}[!ht]
\small
\begin{center}
\caption{Results for Simple Uncertainty Propagation. The best results and second best results are \textbf{bold} and \underline{underlined}, respectively.}
\label{tab:results_sup}
\resizebox{2.0\columnwidth}{!}{\setlength{\tabcolsep}{1.5mm}{
\begin{tabular}{cc|c|c|c|c|c|c|c|c|c}
\toprule
\multicolumn{2}{c|}{}& \multicolumn{3}{c|}{\textbf{HotpotQA}} & \multicolumn{3}{c|}{\textbf{MMLU}}& \multicolumn{3}{c}{\textbf{StrategyQA}} \\
\cmidrule{3-11}
\multicolumn{2}{c|}{\raisebox{2.5ex}[0pt]{\textbf{Method}}}&LLAMA3 8B & LLAMA3 70B & GPT4O & LLAMA3 8B & LLAMA3 70B & GPT4-O & LLAMA3 8B & LLAMA3 70B & GPT4-O \\\midrule
& Arithmetic Mean & 0.695 & 0.676 & 0.781 & 0.621& 0.596 & 0.609&0.576 &0.611 & 0.711\\
& Geometric Mean & 0.713 & 0.714 & \textbf{0.785} & 0.614& 0.591 & 0.610&\underline{0.601} &0.627 & 0.714\\
& RMS & \underline{0.717} & \underline{0.728} & 0.782 & \underline{0.624}& \underline{0.615} & \underline{0.612}&0.584 &\underline{0.629} & \underline{0.723}\\
\midrule
& \textbf{SAUP-Learned} & \textbf{0.771} & \textbf{0.755} & \underline{0.778}  & \textbf{0.669} &\textbf{0.638} & \textbf{0.626}&\textbf{0.787} &\textbf{0.783} & \textbf{0.809}\\
\bottomrule
\end{tabular}}}
\end{center}
\vspace{-1.5em}
\end{table*}

\begin{figure}[ht]
\centering
\includegraphics[width=0.5\textwidth]{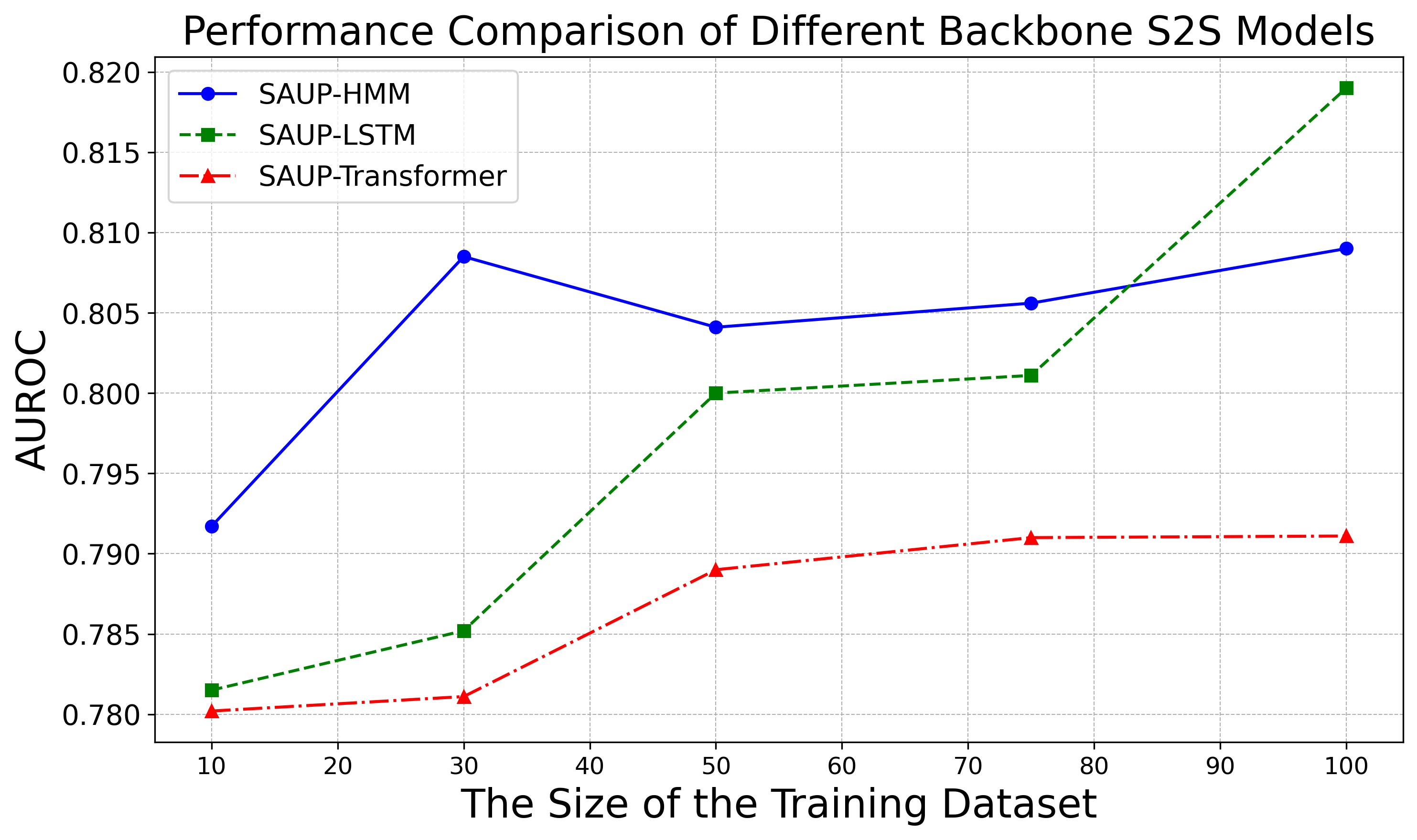}
\caption{The Performance Comparison of Learned-based Surrogates with Various S2S Backbone Models}
\label{fig:Surrogate S2S}
\end{figure}

\subsection{In-Depth Dissection of SAUP}
\label{sec: Dissection}

Given the superiority of our proposed \textit{SAUP}, we further dissect its performance by addressing the following questions. This analysis highlights the advantages of SAUP in various aspects and offers insights into its applicability and performance under different conditions.

\textbf{\textit{Q1: Is the uncertainty measurement of the internal steps beneficial for the overall uncertainty measurement of the agent?}} 

\textit{Yes}, measuring uncertainty at each internal step significantly contributes to a more accurate overall uncertainty estimation. By considering intermediate uncertainties, we capture the cumulative effect of uncertainty propagation throughout the interaction process. As shown in Table \ref{tab:results_all}, SAUP-based methods consistently outperform traditional single-step methods in AUROC scores across datasets and models. The internal step uncertainties provide meaningful information that, when aggregated, enhance the overall uncertainty measurement. Even basic uncertainty propagation methods, such as algorithmic averaging or root mean square (RMS), used to aggregate the uncertainty across all steps, have demonstrated significant improvements over single-step baselines, as shown in Table \ref{tab:results_sup}.

\textbf{\textit{Q2: What is the quality of the surrogates, and how do they benefit the overall uncertainty measurement?}}  

\textit{High-quality surrogates} ensure that situational weights accurately reflect each step's impact on the overall uncertainty. We propose the Position Surrogate (SAUP-P), which assigns greater weight to steps closer to the final answer, and the Plain Distance Surrogate (SAUP-D), which uses only the plain distance. The Hybrid Surrogate (SAUP-PD) combines both approaches with a factor for better balance.

As shown in Table \ref{tab:results_surrogate} and Table \ref{tab:results_sup}, different surrogates improve AUROC scores compared to simple uncertainty propagation baselines, which assign equal weights to all steps. In addition, the HMMD-based (learned) surrogate outperforms others by a clear margin, validating its effectiveness in capturing the agent's situational context.

\begin{figure*}[ht]
\centering
\includegraphics[trim=0cm 5.5cm 0cm 7.0cm, clip, width=1.0\textwidth]{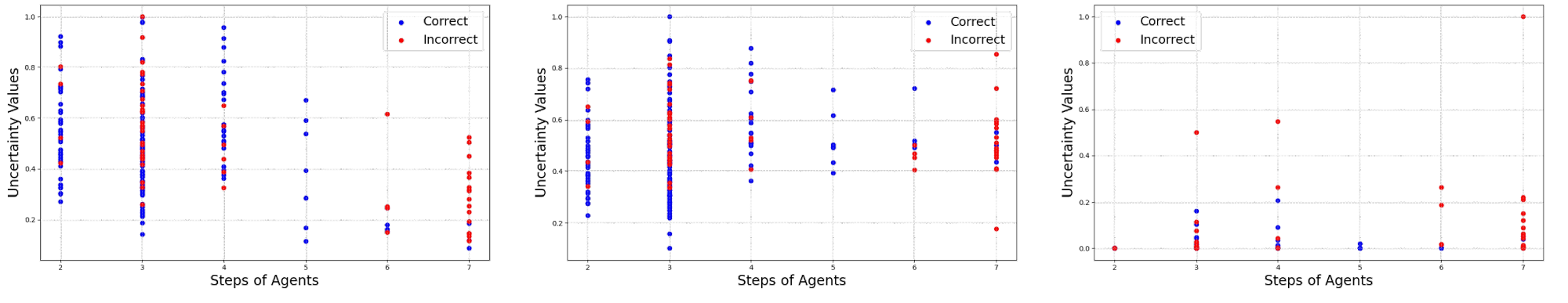}
\caption{Visualization analysis of SAUP on the StrategyQA dataset. Detailed explanations of this figure are provided in the Q3 of Section \ref{sec: Dissection}.}
\label{fig:visual analysis}
\end{figure*}

\textbf{\textit{Q3: Can SAUP demonstrate its superiority in separating correct and incorrect results?}}  

\textit{Yes}, SAUP provides more discriminative uncertainty scores, leading to higher AUROC values across datasets and models, as evidenced in Table \ref{tab:results_all}. The step-by-step propagation of uncertainty allows SAUP to capture the accumulation of uncertainty throughout the reasoning process, enabling better separation of correct and incorrect results.

In addition, we performed a visualization analysis on the StrategyQA dataset (Figure \ref{fig:visual analysis}). The X-axis represents the steps taken, and the Y-axis shows normalized uncertainty values. Red points indicate incorrect answers, and blue points indicate correct answers. SAUP (right sub-image) shows the clearest separation between correct and incorrect answers, outperforming the one-step (left) and simple uncertainty propagation methods (middle), highlighting its advantage in uncertainty estimation.

\textbf{\textit{Q4: Is the HMM reasonable, and how does its performance change with different dataset sizes? Why not use gradient-based models like RNNs or Transformers?}}  

Learned-based surrogates rely on manually annotated data. During training, we map data groups \( D_{a_n} \) and \( D_{o_n} \) to the agent's situational context, enabling SAUP to infer states in unseen scenarios. We use a Hidden Markov Model (HMM) in the main experiment, but also explore LSTM and Transformer models, analyzing their theoretical and experimental advantages.

\textbf{Theoretical Perspective:} HMMs are efficient and interpretable, ideal for limited data but weak in modeling long-range dependencies. LSTMs capture temporal dependencies better but require more data and resources. Transformers handle both local and global dependencies effectively but are computationally expensive and data-intensive.

\textbf{Experimental Comparison:} On the StrategyQA dataset, we evaluated HMM-based, LSTM-based, and mini-size Transformer-based surrogates across varying training dataset sizes. Figure \ref{fig:Surrogate S2S} shows that HMMs perform well with smaller datasets, while LSTMs and Transformers improve with more data. However, Transformer-based surrogates require impractically large datasets for uncertainty measurement tasks, making them less suitable. 

HMMs are practical for uncertainty propagation in LLM-based agents due to their simplicity and efficiency, particularly with limited data. LSTMs are viable alternatives when data and computational resources are sufficient, while Transformers are generally not feasible for most scenarios.

\textbf{\textit{Q5: Does the question difficulty influence the effectiveness of uncertainty propagation?}}  

\textit{Yes}, complex questions lead to longer, nuanced decision-making, increasing uncertainty accumulation. SAUP’s situational awareness framework excels in such cases, effectively propagating uncertainty at each step. As shown in Table \ref{tab:results_all}, SAUP’s advantage is most evident in more challenging datasets like StrategyQA, with greater AUROC improvements.

\section{Conclusion}
In this paper, we propose Situational Awareness Uncertainty Propagation(SAUP), a novel framework for accurately estimating uncertainty in LLM-based multi-step agents. Unlike traditional methods that focus solely on single-step uncertainty, SAUP propagates uncertainty across all steps in the agent's reasoning process and incorporates situational awareness. Experimental results on challenging datasets, show that SAUP outperforms state-of-the-art uncertainty estimation methods, achieving up to 20\% improvements in AUROC scores, thereby demonstrating its effectiveness in enhancing reliability for complex decision-making scenarios. This research highlights the value of multi-step uncertainty estimation and situational awareness in LLM-based agents, providing a strong foundation for their trustworthy deployment.

\section{Limitations}
Despite the effectiveness of SAUP in improving uncertainty estimation for multi-step LLM-based agents, several limitations remain. First, the learning-based surrogate version of SAUP relies on manually annotated datasets for situational weights, which is time-consuming, costly, and may not generalize well to very complex scenarios—especially when manual labels are still prone to errors. Additionally, the complexity of diverse environments could exacerbate the difficulty in ensuring accurate situational labeling. Second, the SAUP framework assumes that uncertainty at each step can be accurately captured. Although this is beyond the scope of our study, errors in single-step uncertainty estimation can compromise the propagation of uncertainty, thereby diminishing the benefits of the SAUP framework. Future work should focus on developing more robust situational weight estimation methods that reduce dependence on manually annotated datasets—potentially leveraging LLM-generated labels—to enhance SAUP's applicability and reliability across diverse use cases.

\bibliography{custom}
\end{document}